\documentclass[11pt]{article}
\usepackage{amssymb}
\usepackage{multicol}
\usepackage{microtype}
\usepackage{indentfirst}
\usepackage{CJKutf8} 
\usepackage{cite} 
\usepackage{amsmath} 
\usepackage{graphicx} 
\usepackage{subcaption}
\usepackage{booktabs}
\usepackage{tabularx}
\usepackage{lineno}
\usepackage{authblk}
\usepackage{hyperref}
\usepackage{xcolor}
\usepackage{soul}        
\usepackage[normalem]{ulem} 

\hypersetup{
    citecolor=green 
}
\usepackage{setspace}
\usepackage{ulem}
\usepackage{natbib}
\usepackage[left=2.5cm, right=2.5cm, top=2.5cm, bottom=2.5cm]{geometry}

\title{An Extended VIIRS-like Artificial Nighttime Light Data Reconstruction (1986-2024)}
\date{}

\author[1, $\dagger$]{Yihe Tian}
\author [2, $\dagger$]{Kwan Man Cheng}
\author[3]{Zhengbo Zhang}
\author[1]{Tao Zhang}
\author[1]{Junning Feng}
\author[4]{Zhehao Ren}
\author[5]{Suju Li}
\author[6]{Dongmei Yan}
\author[1,*]{Bing Xu}

\affil[1]{Department of Earth System Science, Ministry of Education, Ecological Field Station for East Asian Migratory Birds, Tsinghua University, Beijing 100084, China}
\affil[2]{Department of Computer Sciences, University of Wisconsin-Madison, Madison 53703, USA}
\affil[3]{Institute of Automation, Chinese Academy of Sciences, Beijing 100190, China}
\affil[4]{China Association for International Exchange of Personnel, Beijing 100038, China}
\affil[5]{National Disaster Reduction Center of China, Beijing 100124, China}
\affil[6]{Aerospace Information Research Institute, CAS, Beijing 100094, China}

\affil[$\dagger$]{Equal contribution}
\affil[*]{Corresponding author} 

\begin{document}

\begin{CJK*}{UTF8}{gbsn} 
\maketitle

\begin{abstract}
Artificial Night-Time Light (NTL) remote sensing is a vital proxy for quantifying the intensity and spatial distribution of human activities. Although the NPP-VIIRS sensor provides high-quality NTL observations, its temporal coverage, which begins in 2012, restricts long-term time-series studies that extend to earlier periods. Current extended VIIRS-like NTL data products suffer from two significant shortcomings: the underestimation of light intensity and the omission of structural details. To overcome these limitations, we present the Extended VIIRS-like Artificial Nighttime Light (EVAL) dataset, a new annual NTL dataset for China spanning from 1986 to 2024. This dataset was generated using a novel two-stage deep learning model designed to address the aforementioned shortcomings. The model first constructs an initial estimate and subsequently refines fine-grained structural details using high-resolution impervious surface data as guidance. Quantitative evaluations demonstrate that EVAL significantly outperforms state-of-the-art products, exhibiting superior temporal consistency and a stronger correlation with socioeconomic indicators.

\end{abstract}

\section{Background \& Summary}
Remote sensing of Night-Time Light (NTL) offers a unique modality for Earth observation by capturing the radiant characteristics of light sources originating from nocturnal human activities. This provides a critical dimension of information unattainable through conventional daytime remote sensing \citep{elvidge1997mapping,zheng2023nighttime,levin2020remote}. Unlike daytime remote sensing, which primarily relies on reflected solar radiation, NTL data directly record the intensity and distribution of artificial light, establishing a strong proxy for human presence and socioeconomic dynamics. With the accelerating pace of global urbanization and the expansion of human activities, NTL data have demonstrated significant scientific value in diverse fields such as land use change monitoring \citep{chen2022global,ma2023global,zhang2011mapping}, socioeconomic development assessment \citep{bennett2017advances,sutton1997modeling,ma2014responses}, urbanization process analysis \citep{elvidge1997mapping,henderson2003validation,liu2024enhancing}, carbon emission estimation \citep{jung2022does,fang2022drives,xu2023spatio}, and ecological environment change monitoring \citep{ma2023global,zhang2023spatial,gaston2013ecological}. Furthermore, NTL remote sensing plays a crucial role in advancing the United Nations Sustainable Development Goals (SDGs). It not only enables the direct assessment of progress towards SDG 7.1 (Affordable and Clean Energy) and SDG 11.1 (Sustainable Cities and Communities)\citep{lv2025advancing,levin2025challenges}, but also indirectly sheds light on other human activity-related goals, including SDG 1.1, SDG 1.2, SDG 8.1, and SDG 10.2\citep{zhang2023spatial}.

Currently, NTL remote sensing data are primarily sourced from two major satellite systems: the Operational Linescan System (OLS) on the Defense Meteorological Satellite Program (DMSP), and the Visible Infrared Imaging Radiometer Suite (VIIRS) aboard the Suomi National Polar-orbiting Partnership (NPP). Specifically, DMSP-OLS NTL data cover a continuous observational record from 1992 to 2013, offering a relatively complete history of nighttime light variations\citep{wu2021developing}. In contrast, NPP-VIIRS NTL data, operational since 2012, boast superior spatial resolution and radiometric sensitivity\citep{shi2014evaluation}. However, these two systems employ different measurement principles, radiometric calibration methods, and data quantization units, resulting in data formats and numerical ranges that lack direct comparability. This heterogeneity significantly limits their applicability in long-term time-series analyses. Without specialized cross-platform conversion, integrated use of these datasets is infeasible, thereby restricting the scope and depth of long-term studies based on NTL data\citep{levin2020remote}.

In response to these challenges, scholars in recent years have dedicated efforts to developing various data conversion and calibration methods to construct unified, long-term NTL datasets. Mainstream integration methods often use DMSP-OLS as the baseline. Zhang et al. \citep{zhang2016robust} used ridge regression for calibration, while Li et al. \citep{li2020harmonized} applied kernel density estimation to convert NPP-VIIRS data into a DMSP-OLS-like product. Similarly, Wu et al. \citep{wu2021developing} adopted a pseudo-invariant pixel method. however, such approaches inherit the inherent deficiencies of the original DMSP-OLS data\citep{levin2020remote}, such as coarse spatial resolution and pixel over-saturation, which severely compromises the accuracy and reliability of the resulting data. Subsequent research targeted these specific issues: Cao et al. \citep{cao2019simple} utilized a self-adjusting model to mitigate over-saturation, while Yu et al. \citep{yu2025reconnecting} developed the 500-meter Tongji-NTL dataset through a super-resolution framework. Despite these improvements, a critical limitation persists: these datasets are not calibrated to the absolute radiometric units of NPP-VIIRS, which constrains their physical interpretation and quantitative use. This gap points toward a more robust paradigm: upgrading historical DMSP data to match the superior standard of NPP-VIIRS, which excels in both spatial detail and radiometric precision. In this frontier, Chen et al.\citep{chen2020extended} proposed a cross-sensor calibration method, introducing the Enhanced Vegetation Index (EVI) as an auxiliary variable to augment the original DMSP NTL data into an Enhanced Artificial Nighttime Light Index (EANTLI), partially overcoming the issue of information loss in DMSP data. Subsequently, their research combined an autoencoder model to convert DMSP-OLS NTL data into an NPP-VIIRS-like format, generating a global NPP-VIIRS-like NTL dataset for the period of 2000 to 2018 named LongNTL. Building upon this, Chen et al.\citep{chen2024global} further refined the conversion method by integrating original DMSP-OLS NTL data with annual Normalized Difference Vegetation Index (NDVI) data. They introduced a U-Net deep learning model as the core architecture for the conversion, which preserves both global semantic information and local detail features, ultimately producing a global Simulated VIIRS Nightime Light Dataset (SVNL) covering the years 1992 to 2023.

Although previous studies have established initial cross-sensor mappings from DMSP-OLS to NPP-VIIRS NTL data, enabling long-term integration, several technical limitations remain such as underestimation of light intensity and structural omission. First, existing methods underperform in regions with high radiance values—such as central business districts, industrial complexes, and ports—due to the saturation inherent in DMSP-OLS data. In these areas, the loss of gradient information severely hampers the model's ability to reconstruct light intensity variations, resulting in distorted urban representations and biased estimates of socioeconomic indicators. Second, current VIIRS-like datasets omit intra-urban structures and road networks, limiting their effectiveness for analyzing regional development and transportation infrastructure. This shortcoming arises primarily from the auxiliary features employed—namely, spectral indices like EVI and NDVI—that are optimized for vegetated surfaces. These indices exhibit low sensitivity in non-vegetated areas such as urban built-up zones, industrial regions, transportation corridors, and arid landscapes\citep{small2001estimation}, making it difficult to distinguish between anthropogenic and natural barren surfaces. As a result, these models struggle to reconstruct fine-scale urban structures and road connectivity. Moreover, most prior approaches rely on generic Autoencoder\citep{ng2011sparse}  or U-Net\citep{ronneberger2015u}. These methods primarily employ generic architectures that are not optimized for the unique challenges of NTL data, such as its extreme dynamic range and pixel saturation artifacts. Unlike standard daytime optical imagery, NTL data requires specialized handling to resolve these issues. Consequently, current models often struggle to accurately reconstruct high light intensities and fine spatial details.

In summary, although existing approaches for generating NPP-VIIRS-like NTL data have achieved notable progress, they continue to suffer from two key deficiencies: intensity underestimation and structural omission. These shortcomings limit their suitability for long-term time-series analyses. To overcome these challenges, this study introduces a novel two-stage deep learning framework specifically designed to produce high-quality NPP-VIIRS-like NTL data. The framework first performs an initial reconstruction and subsequently refines fine-grained structural details by fusing multi-scale information and leveraging high-resolution impervious surface masks for guidance. Using this methodology, we constructed the Extended VIIRS-like Artificial Nighttime Light (EVAL) dataset. It provides a high-quality, annual, 500m resolution, NPP-VIIRS-like NTL time series for China, spanning the period from 1986 to 2024. This dataset effectively addresses the problems of underestimation and structural omission found in existing data and provides continuous and consistent data support for related research endeavors.

\section{Methods}

\subsection{Data sources}
This study primarily encompasses two categories of data sources as shown in Table. \ref{tb1}: Night-Time Light remote sensing data and auxiliary geospatial data. NTL data serves as a critical indicator for characterizing the intensity and spatial distribution of human activities, forming the foundation of the reconstruction model. Auxiliary data provide finer surface characteristics to address the inherent differences in spatial resolution, radiometric detection range, and on-orbit calibration between the DMSP-OLS and NPP-VIIRS sensors, thereby enhancing the precision and reliability of the reconstruction model.

\paragraph{DMSP-OLS NTL}

The original DMSP-OLS NTL data suffer from temporal inconsistencies and fluctuations in data quality due to several issues, including the lack of on-board calibration, inter-sensor performance discrepancies, and sensor degradation over time. To overcome these limitations, this study utilizes the systematically, stepwise-calibrated global DMSP-OLS NTL dataset developed by Li et al.\citep{li2020harmoniz} (\url{https://doi.org/10.6084/m9.figshare.9828827.v2}). This dataset exhibits enhanced temporal consistency and comparability. Consequently, it was selected as the baseline DMSP-like data for the period of 1992 to 2013, serving as the foundation for both the training and inference processes of our model.

To further extend the temporal scope of the research and cover periods where DMSP data is unavailable, this study incorporates the PANDA-China dataset\citep{zhang2024prolonged} as a supplement (\url{https://data.tpdc.ac.cn/en/data/e755f1ba-9cd1-4e43-98ca-cd081b5a0b3e/}). In this dataset, PANDA-China provide a high-precision and temporally consistent DMSP-like NTL data product spanning from 1984 to 2020. In this research, the PANDA-China dataset was employed for model inference during 1986-1991, the specific intervals where the original DMSP data is absent..

Considering the inherent dissimilarities between the PANDA-China and Li's NTL datasets, we trained a simple U-Net model\citep{ronneberger2015u} for the year 1992. This model was used to map the PANDA-China NTL data to the characteristics of Li's NTL data for the 1986-1991period.

\paragraph{NPP-VIIRS NTL}
This study utilizes the annual mean composite NTL dataset from the NPP-VIIRS Day/Night Band\citep{elvidge2021annual} for the years 2012 to 2024 (\url{https://eogdata.mines.edu/nighttime_light/annual/v20/}), released by the Earth Observation Group at the Colorado School of Mines. This dataset is generated from monthly NPP-VIIRS data, during which interferences from ephemeral or anomalous light signals—such as those from fires, auroras, high-energy particle impacts, and background noise—are systematically removed in the synthesis process. To address the issue of missing data in parts of Northern China for several months in 2012, this study adheres to the outlier removal and annual mean calculation methodology proposed by Elvidge et al.\citep{elvidge2021annual} to ensure consistency between the data for that year and subsequent years.

The research proceeds on the assumption that major urban centers exhibit the highest NTL intensity. A pixel is identified as an anomaly and adjusted if its NTL intensity surpasses the maximum observed in major cities. In the course of this study, four of the most developed representative cities in China were selected: Beijing, Shanghai, Hong Kong, and Taipei. The average of the maximum NTL intensity values from these four cities was established as a threshold. Any NPP-VIIRS NTL value exceeding this threshold was replaced with the average NTL intensity of its 24 neighboring pixels.
According to the findings of Zhao et al.\citep{zhao2019building}, the distribution of NPP-VIIRS NTL data after a logarithmic transformation more closely resembles that of DMSP-OLS NTL data. Therefore, during the model training and prediction phases, we used the natural log-transformed NPP-VIIRS NTL data as the target variable. Upon completion of the prediction, an exponential transformation was applied to map the generated results back to the original data space, thereby ensuring the output is consistent with the actual data.

\paragraph{Landsat surface reflectance}
To acquire fine-scale land cover information, this study utilized the Tier 1 Surface Reflectance (SR) products from the Landsat series of satellites (TM, ETM+, and OLI)\citep{arvidson2001landsat,wulder2019current}. The data were sourced from the United States Geological Survey (USGS) and included six spectral bands\citep{loveland2016landsat}: Blue, Green, Red, Near-Infrared (NIR), Shortwave Infrared 1 (SWIR1), and Shortwave Infrared 2 (SWIR2). These SR products have undergone atmospheric correction using the LEDAPS\citep{masek2006landsat} and LaSRC\citep{vermote2016preliminary} algorithms employed by the USGS, which mitigates the impact of atmospheric effects on the surface reflectance signal.
To further minimize the influence of clouds, cloud shadows, and snow/ice cover, this study computed the 10th percentile of the surface reflectance values for each spectral band from all available observations within a given year, which was then used as the representative annual surface reflectance. Furthermore, to ensure temporal consistency across data from different sensors, the SR data from the OLI sensor were normalized to the level of the TM and ETM+ sensors. This harmonization was based on the transformation coefficients proposed by Roy et al.\citep{roy2016characterization}, thereby guaranteeing the comparability of the multi-temporal Landsat imagery across different years.
\paragraph{Artificial impervious surface}

To accurately capture the structure of the built environment within cities and the connectivity features between regions, this study incorporated the Global Artificial Impervious Areas (GAIA) dataset\citep{gong2020annual} (\url{https://data-starcloud.pcl.ac.cn/zh/resource/13}). GAIA provides annual 30-meter spatial resolution masks of the Earth's impervious surfaces covering the entire study period. This dataset is characterized by a global overall accuracy exceeding 90\% and exhibits good time-series consistency.

\begin{table}[htbp]
  \centering 
  \caption{Description of datasets used} 
  \label{tb1}
  \begin{tabularx}{\textwidth}{
    X 
    >{\raggedright\arraybackslash}X 
    >{\raggedright\arraybackslash}X}
    \toprule
    \textbf{Dataset} & \textbf{Description} & \textbf{Data Source} \\
    \midrule
    Li's Nighttime Light Dataset\citep{li2020harmonized} & Served as DMSP-OLS NTL features for training and inference (1992-2013). & \url{https://doi.org/10.6084/m9.figshare.9828827.v2} \\
    
    PANDA-China\citep{zhang2021prolonged} & Served as DMSP-OLS NTL features for early-period time-serious (1986-1991). & \url{https://doi.org/10.11888/Socioeco.tpdc.271202} \\
    
    Annual VNL V2\citep{elvidge2021annual} & Served as the ground truth NPP-VIIRS NTL data for model training and accuracy assessment. & \url{https://eogdata.mines.edu/nighttime_light/annual/v20/} \\
    
    Landsat Surface Reflectance & Used six spectral bands (Blue, Green, Red, NIR, SWIR1, SWIR2) as features for training and inference. & Google Earth Engine \\
    
    Global Artificial Impervious Area\citep{gong2020annual} & Served as the impervious surface area mask to guide model training and inference. & \url{https://data-starcloud.pcl.ac.cn/zh/resource/13} \\
    
    LongNTL\citep{chen2020global} & Used as a benchmark dataset for accuracy assessment. & \url{https://doi.org/10.7910/DVN/YGIVCD} \\
    
    SVNL\citep{chen2023history} & Used as a benchmark dataset for accuracy assessment. & \url{https://doi.org/10.6084/m9.figshare.22262545.v8} \\
    
    Resident Population of Provinces in China & Used for correlation analysis with the generated NTL data. & China Statistical Yearbook (1986-2020) \\
    
    Gross Domestic Product of Provinces in China & Used for correlation analysis with the generated NTL data. & China Statistical Yearbook (1986-2020) \\
    \bottomrule
  \end{tabularx}
\end{table}
\subsection{Framework}

For the task of reconstructing NPP-VIIRS-like NTL data, we propose a framework, upon which we design the \textbf{Hierarchical Fusion Decoder (HFD)} and the \textbf{Dual Feature Refiner (DFR)}. Within this framework, the encoder, decoder, and refiner components are modular and can be readily replaced to potentially achieve improved performance.

\paragraph{Pipeline}

\begin{figure}[!htbp] 
  \centering
  \includegraphics[width=0.6\textwidth]{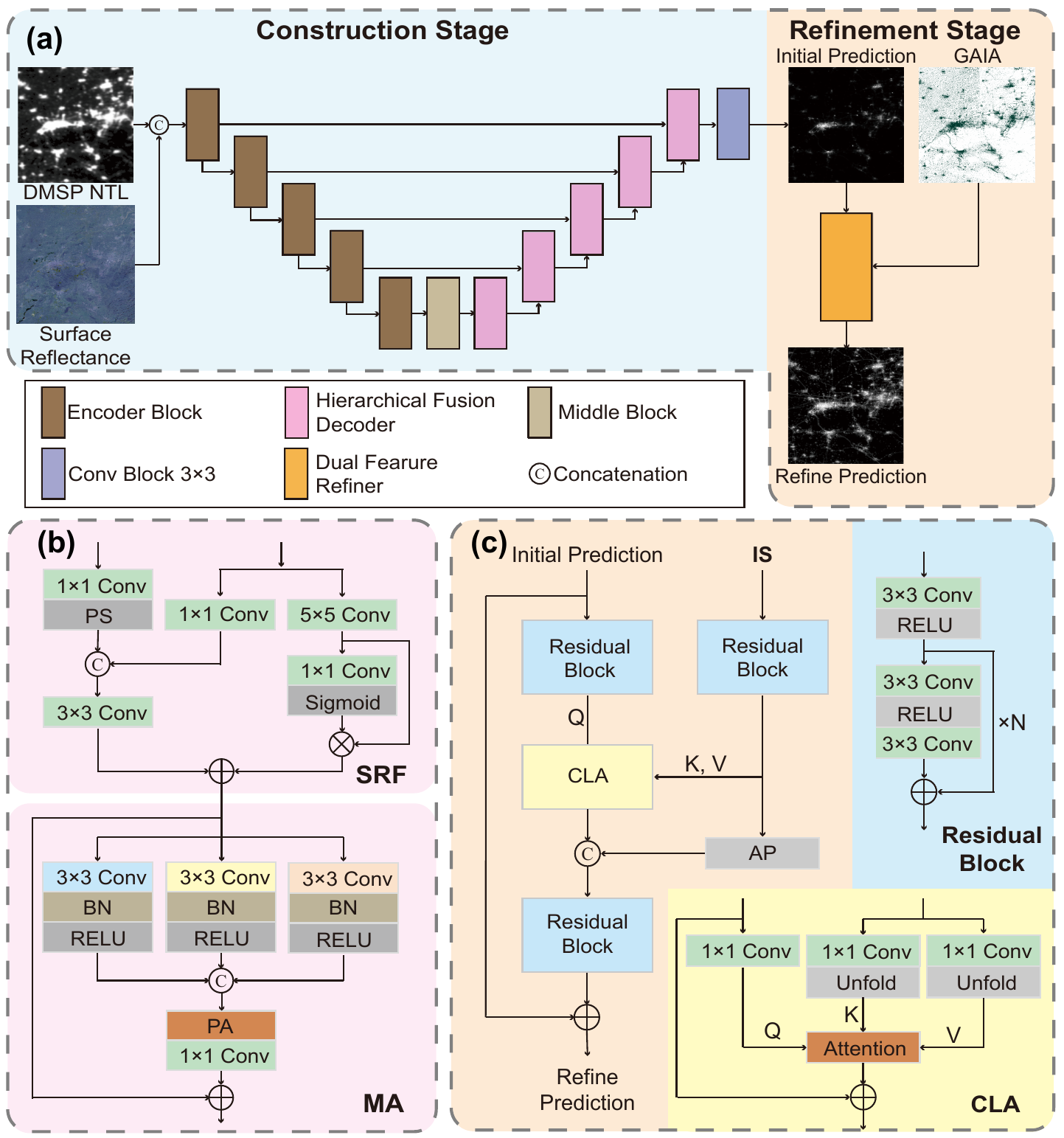}
  \caption{The architecture of our proposed framework, detailing the overall pipeline and its key components. \textbf{(a)} The overall pipeline consists of a construction stage and a refinement stage. The construction stage uses a U-Net-based architecture to generate an initial prediction via our specialized decoder (HFD). The subsequent refinement stage then performs fine-grained adjustments on this prediction using the refiner module (DFR). \textbf{(b)} The detailed structure of the HFD. It is composed of a Structure Residual Fusion (SRF) module, which intelligently incorporates fine-grained details from the encoder's skip connection, and a Multi-scale Aggregator (MA), which captures and adaptively fuses contextual features from varying receptive fields. \textbf{(c)} The detailed structure of the DFR module. It utilizes residual blocks and a Cross-Resolution Local Attention to fine-tune the reconstructed image, guided by high-resolution features. }
  \label{fig1}
\end{figure}

The overall architecture of our proposed framework is illustrated in Figure. \ref{fig1} and consists of a backbone network followed by an attached refiner module. The backbone model adopts a U-Net\citep{ronneberger2015u} architecture, comprising an encoder, a decoder, and skip connections. It accepts DMSP NTL and surface reflectance data as input to reconstruct the log-transformed NPP-VIIRS NTL data.

Within the backbone, a standard visual encoder produces multi-scale features through five downsampling stages, mirrored by five upsampling stages in the decoder that fuse encoder and decoder features via skip connections. To better address the challenges of nighttime light reconstruction, we introduce the HFD comprising a Structure Residual Fusion (SRF) module and a Multiscale Aggregator (MA). The SRF enhances structural detail and edge sharpness by refining skip features while suppressing mis-registered noise. The MA, in parallel, adaptively fuses multi-scale contextual cues to reduce over-spread artifacts. Together, they yield clearer and more consistent reconstructions.

Following the backbone network, we separately train the DFR. This module utilizes a Cross-Resolution Local Attention, leveraging high-resolution impervious surface area masks as guidance to optimize the expression of fine-grained structural features, which greatly improves the results for internal urban fabric and road networks.

The model employs the Mean Squared Error (MSE) loss function during the training of the construction stage:
\begin{equation} \label{eq:mse_loss}
    \mathcal{L}_{MSE} = \frac{1}{n} \sum_{i=1}^{n} (y_i - \hat{y}_i)^2
\end{equation}
\(y_i\) is the true target value for the \(i\)-th sample, and \(\hat{y}_i\) is the predicted value generated by the model for the \(i\)-th sample.

The L1 loss (Mean Absolute Error) is used for training the refinement stage:
\begin{equation} \label{eq:l1_loss}
    \mathcal{L}_{L1} = \frac{1}{n} \sum_{i=1}^{n} |y_i - \hat{y}_i|
\end{equation}
\(y_i\) is the true target value for the \(i\)-th sample, and \(\hat{y}_i\) is the predicted value output by the model for the \(i\)-th sample.

\paragraph{Hierarchical fusion decoder}

As shown in Figure~\ref{fig1} (b), the HFD processes skip-connection features and upsampled features from the previous layer, and is composed of an SRF and an MA module. The SRF module processes skip-connections in parallel: a 1x1 convolution for "content" and a 5x5 convolution for "structure." The "content" is concatenated with the upsampled features, while the "structure" information is added as a residual, enhancing edge and texture recovery. The MA module, inspired by DarkIR\citep{feijoo2025darkir}, then uses three parallel dilated convolution branches to capture multi-scale contextual information. A lightweight channel attention mechanism adaptively weights and sums the features from these branches.

\paragraph{Dual feature refiner}

The DFR, illustrated in Figure~\ref{fig1} (c), is a cross-resolution module that performs localized, fine-grained corrections. It leverages high-resolution impervious surface area masks to guide the refinement of the initial NTL prediction by outputting a residual correction. The DFR employs a dual-branch architecture using residual blocks to extract features. Information is fused via a Cross-Resolution Local Attention mechanism, where each low-resolution pixel attends to a 5x5 local neighborhood in the high-resolution feature map using scaled dot-product attention. This allows for the adaptive extraction of relevant texture details. The enhanced low-resolution features are then fused with downsampled high-resolution features and passed through final residual blocks to predict the residual correction.

\subsection{Evaluation metrics}
To conduct a quantitative evaluation, we selected four authoritative metrics: the Coefficient of Determination ($\text{R}^2$), Root Mean Square Error (RMSE), Peak Signal-to-Noise Ratio (PSNR)\citep{hore2010image}, and the Universal Image Quality Index (UIQI)\citep{wang2002universal}.

\subsection{Implementation details}
The model was implemented in Python using PyTorch and trained on a single NVIDIA RTX 3090 Ti GPU. It utilized 20,000 512x512 pixel image patches from a 2013 China-wide dataset, containing co-registered night-time light, Landsat, and impervious surface data. To address the data imbalance from large dark areas, these patches were selected only if at least 1\% of their pixels had non-zero values. 
These patches were split into 80/20 for training and validation. Input data was resampled to 250m, while the network architecture featured a ResNet-50 backbone, a symmetric decoder with the proposed HFD module, and a Dual-Feature Refiner (DFR) to enhance structural details. The optimization was a two-stage process: the main network was trained for 60 epochs using the Adam optimizer with a learning rate of 1e-4, followed by an additional 10 epochs of training for only the DFR module using L1 loss and a lower learning rate of 1e-5. To generate the final nationwide imagery, we used a sliding window approach with a 6-pixel overlap between adjacent patches. The average value in these overlapping regions was used as the final prediction to ensure a seamless and consistent reconstruction.

\section{Data Records}
The complete EVAL dataset, covering the full period from 1986 to 2024, is publicly available at the National Tibetan Plateau Data Center (TPDC) (\url{https://doi.org/10.11888/HumanNat.tpdc.302930})\cite{tian2025extended}. The EVAL dataset is a composite product designed to provide a continuous, long-term time series. It consists of two distinct components corresponding to different periods:

\textbf{Reconstructed Period (1986-2013)}: This part of the dataset contains the annual, 500m resolution VIIRS-like NTL data generated by our proposed reconstruction framework. To produce this, we used two different DMSP-OLS inputs: for the 1992-2013 period, we used Li's NTL; for the 1986-1991 period, we used PANDA-China. The reliability of the data from this earlier 1986-1991 period is specifically addressed in the "Validation of early-period time-series" section.

\textbf{Processed Official Period (2012-2024)}: This part consists of the official annual mean NPP-VIIRS DNB data, which we have processed by capping high-value pixels to ensure consistency.

The overlapping years of 2012 and 2013 serve a critical purpose: they allow for a direct comparison between our reconstructed data and the official data, enabling a robust evaluation and validation of our model's performance.

\section{Technical Validation} 
\begin{figure}[h] 
  \centering
  \includegraphics[width=0.8\textwidth]{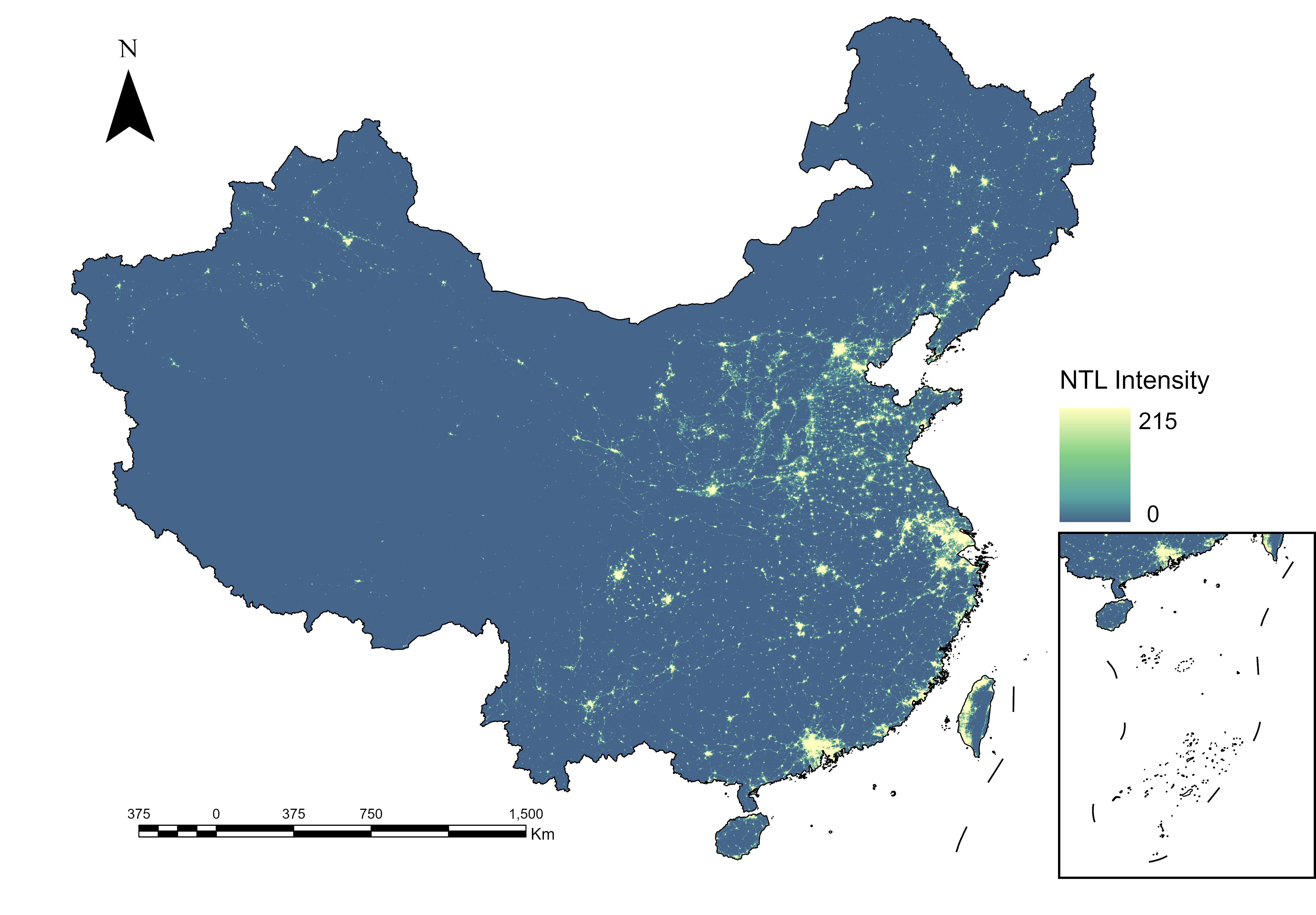}
  \caption{The EVAL product for the year 2012. The image has been contrast-stretched using histogram equalization for visualization purposes.}
  \label{fig2}
\end{figure}

Based on the framework and modules proposed in this study, we generated EVAL: an annual, 500m resolution, NPP-VIIRS-like nighttime light dataset for China, with coverage beginning in 1986, as illustrated in Figure. \ref{fig2}. To the best of our knowledge, EVAL represents the VIIRS-like NTL product with the longest available time series to date. In this section, we will assess the performance of the EVAL dataset from various perspectives. 

\subsection{Accuracy assessment}
Considering that the synchronous observation window of the DMSP-OLS and NPP-VIIRS sensors is limited to 2012–2013, and that the samples used for model training in this study were sourced from 2013, we selected the entire region of China in 2012 as a spatiotemporally independent test area. 

We first compared our EVAL dataset against two existing products, LongNTL and SVNL, at the pixel scale. The results, summarized in Table. \ref{tb2}, show that EVAL achieved superior performance across all evaluation metrics.
In terms of model goodness-of-fit, the R² of EVAL reached 0.8088, which is significantly higher than that of SVNL (R² = 0.6857) and LongNTL (R² = 0.5961). Regarding prediction accuracy, EVAL achieved a RMSE of 0.9965, representing a reduction of 0.2811 and 0.4518 compared to SVNL (RMSE = 1.2776) and LongNTL (RMSE = 1.4483), respectively. For the PSNR, EVAL (PSNR = 46.6431db) also significantly surpassed SVNL (PSNR = 44.5936db) and LongNTL (PSNR = 43.5045db), further confirming the high degree of consistency between its predicted and observed values
Furthermore, with respect to the UIQI, EVAL (UIQI = 0.8962) exhibited the most prominent performance, markedly outperforming SVNL (UIQI = 0.7726) and LongNTL (UIQI = 0.6796). This result underscores its comprehensive advantages in preserving luminance, contrast, and structure.
\begin{table}[h]
\centering
\caption{\textbf{Quantitative evaluation in 2012.} EVAL achieved state-of-the-art results across all metrics.}
\label{tb2}
\begin{tabular}{c|cccc}
\hline
Products & $\text{R}^2$\textbf{$\uparrow$} & RMSE\textbf{$\downarrow$} & PSNR\textbf{$\uparrow$} & UIQI\textbf{$\uparrow$}\\
\hline
LongNTL\citep{chen2020extended} & 0.5961 & 1.4483 & 43.5045 & 0.6796\\
SVNL\citep{chen2024global} & 0.6857 & 1.2776 & 44.5936 & 0.7726\\
\textbf{EVAL} &\textbf{0.8088} & \textbf{0.9965} & \textbf{46.7525} & \textbf{0.8968}\\
\hline
\end{tabular}
\end{table}

Following the pixel-level evaluation, we assessed the accuracy of our EVAL dataset at the city scale. This analysis used 2891 county-level administrative units across China. For each unit, we calculated the total nighttime light intensity for all three NTL products. We then compared these aggregate values against the ground truth data using the Coefficient of Determination and RMSE.

The results show that EVAL also performs best at this broader scale. It achieved an R² of 0.975, which is 7.7\% higher than SVNL (0.905) and 21.9\% higher than LongNTL (0.800). Similarly, its RMSE of 1394.43 was 48.6\% and 64.7\% lower than the two comparative methods, respectively.

\begin{figure}[h] 
  \centering
  \includegraphics[width=1\textwidth]{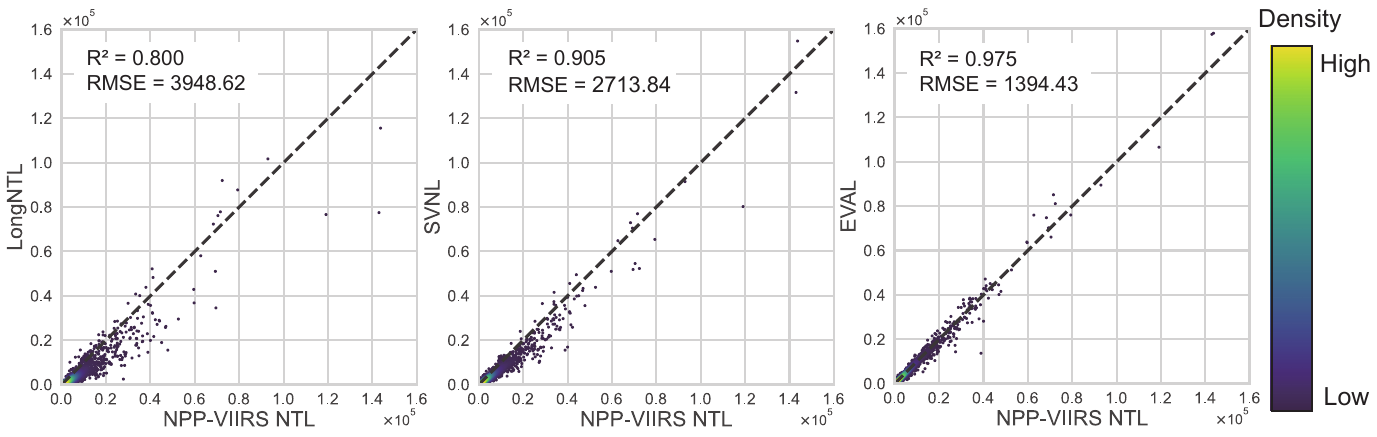}
  \caption{The scatter plot comparing VIIRS-like NTL data with actual NPP-VIIRS NTL data at the city scale.}
  \label{fig3}
\end{figure}

The scatter plot in Figure. \ref{fig3} visualizes these findings. For LongNTL and SVNL, many data points fall below the 1:1 line, revealing a systematic underestimation. In contrast, the points for our EVAL dataset adhere closely to the 1:1 line across the entire data range. This demonstrates that our method effectively resolves the underestimation problem at the regional scale.

\begin{figure}[h] 
  \centering
  \includegraphics[width=0.7\textwidth]{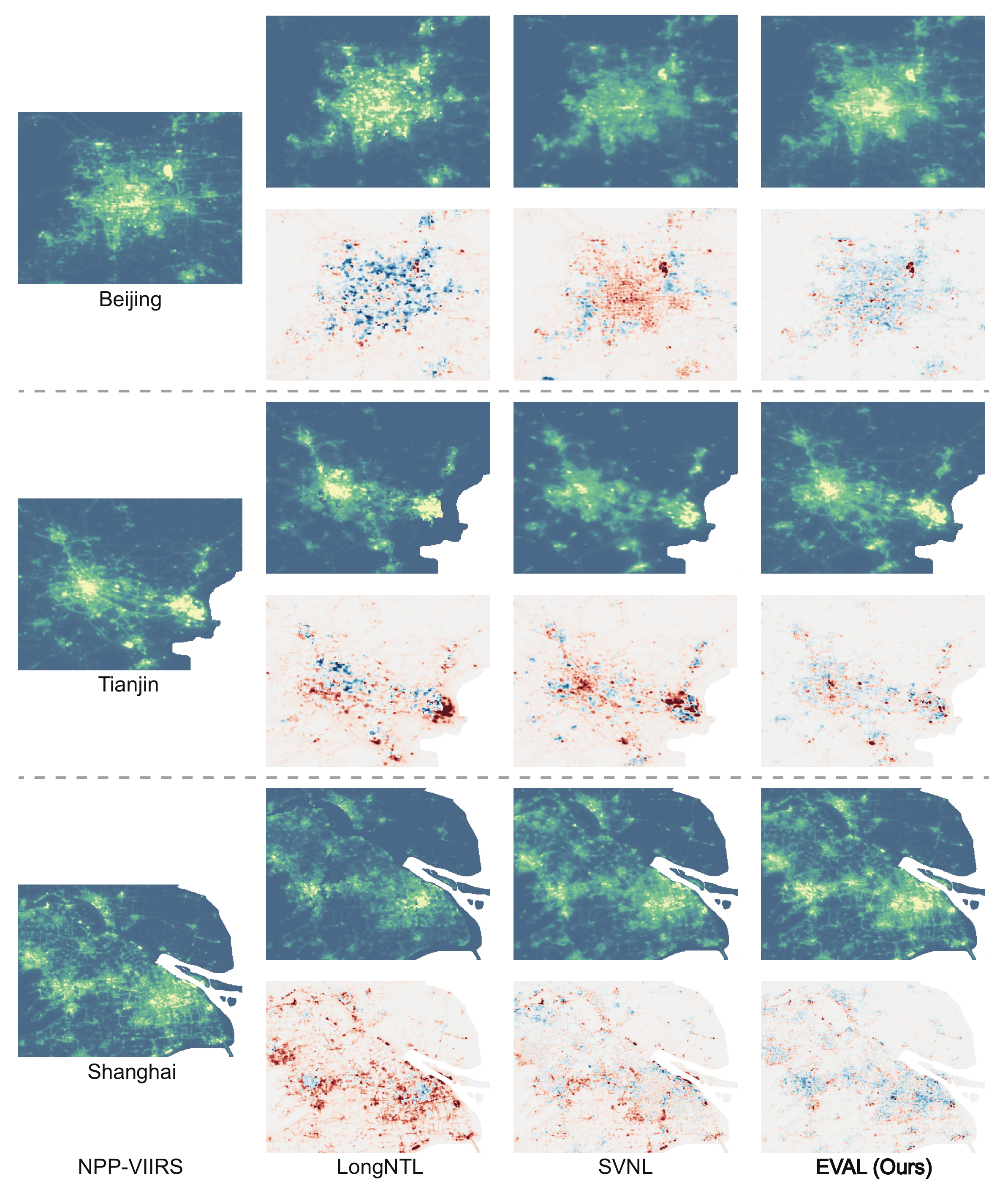}
  \caption{Analysis of prediction results in major urban areas for the year 2012, specifically the megacities of Beijing, Tianjin, and Shanghai. Blue and red denote overestimation and underestimation, respectively, where color intensity is proportional to the magnitude of the deviation.}
  \label{fig4}
\end{figure}

\begin{figure}[h] 
  \centering
  \includegraphics[width=0.7\textwidth]{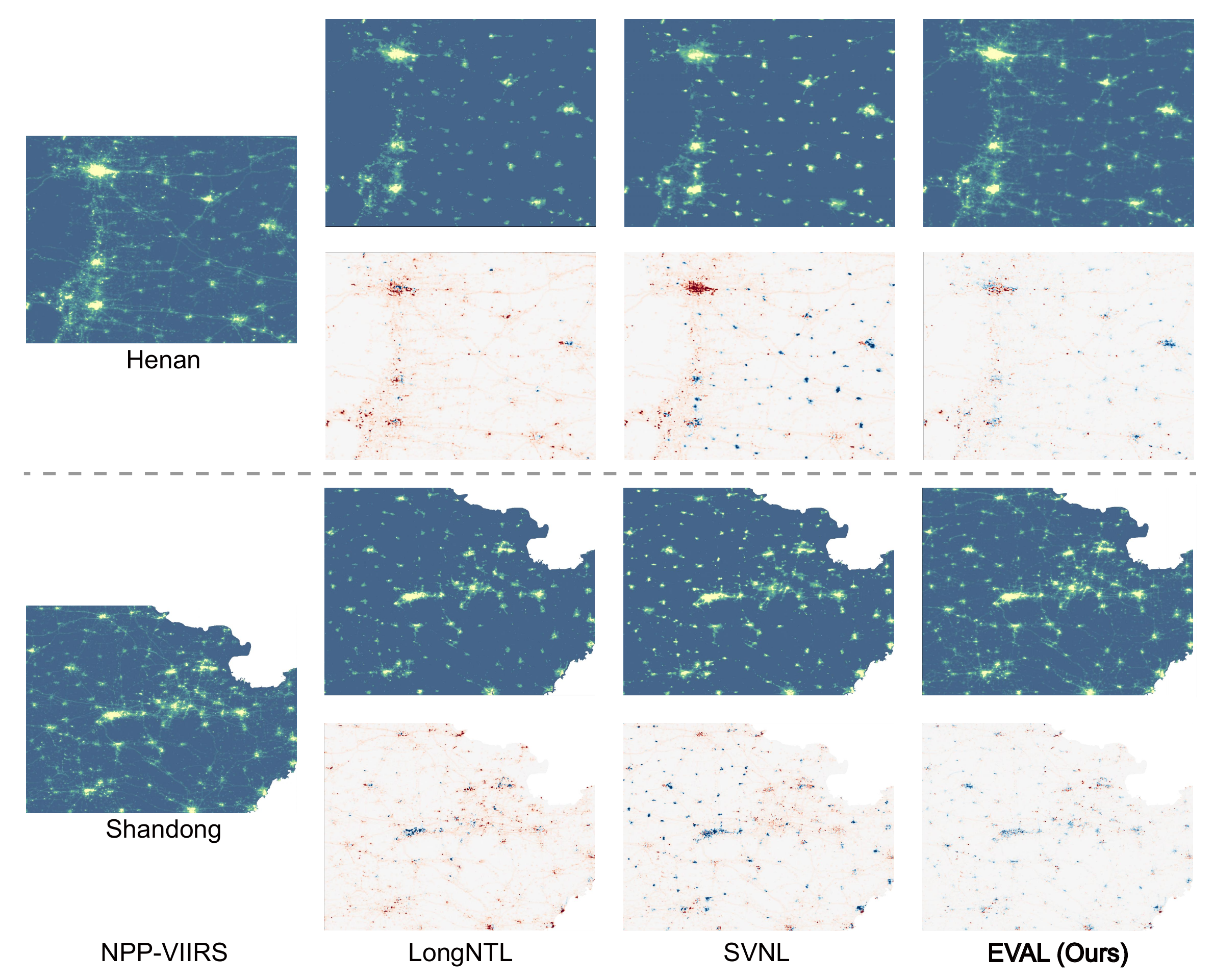}
  \caption{Analysis of prediction results in rural areas and along road networks for the year 2012, focusing on Henan and Shandong—two provinces characterized by dense rural settlements. Blue and red denote overestimation and underestimation, respectively, where color intensity is proportional to the magnitude of the deviation.}
  \label{fig5}
\end{figure}

The overall superiority of the EVAL at the regional scale can be primarily attributed to its effective resolution of the underestimation and omission issues that are prevalent in existing VIIRS-like NTL datasets. A detailed comparative analysis of local areas further substantiates the exceptional performance and robustness of EVAL across different regions. In highly economically developed urban areas with extremely intense nighttime light shown in Figure. \ref{fig4}. The results reveal that LongNTL produces volatile and large-magnitude estimation errors within urban areas, whereas SVNL fails to capture gradients in light intensity, leading to a systematic underestimation across the region. EVAL not only corrects the severe underestimation present in LongNTL and SVNL but also reconstructs the spatial gradients of the nighttime light distribution. Meanwhile, in extensive rural areas and across road networks like Figure. \ref{fig5} shows, both LongNTL and SVNL fail to reconstruct the road networks. Furthermore, SVNL tends to systematically overestimate the nighttime light intensity in smaller settlements. EVAL demonstrates significantly higher completeness and finer spatial detail compared to LongNTL and SVNL. This outcome clearly demonstrates that the method proposed in this study enhances the capacity of the generated data to represent fine spatial detail while simultaneously and significantly improving its predictive accuracy in high-brightness regions.

\subsection{Validation of early-period time-series}
We generated the EVAL dataset in two distinct phases. For the 1992–2013 period, we used the Li's Nighttime Light dataset as input, whose reliability has been established in prior tests. In contrast, for the previous period from 1986 to 1991, we used the PANDA-China dataset as input, which required additional validation to confirm its robustness. 

To perform this validation, we designed a spatial stability evaluation using GDP and Resident Population (POP) data as proxies for human activity. The core assumption is that the spatial pattern of the relationship between NTL and these proxies should be very stable over time. Significant data quality issues, like sensor noise or calibration errors, would noticeably disrupt this stable spatial pattern. 

Our method proceeds in two steps.

\textbf{Normalization } To remove the effect of overall economic growth, we compute a normalized statistic for each province $i$ in each year $j$:
    \begin{equation} \label{eq:spatial_norm}
        n_{ij} = \frac{\mathrm{NTL}_{ij}}{K_{ij}},
    \end{equation}
where $\mathrm{NTL}_{ij}$ is the nighttime light intensity for province $i$ in year $j$, and $K_{ij}$ is an external proxy for the economic scale of that province in that year. Thus, $n_{ij}$ represents nighttime light intensity per unit of the proxy.
We arrange all $n_{ij}$ values into a matrix $N = [n_{ij}]$, where each column
    \[
        N_j = (n_{1j}, n_{2j}, \ldots, n_{Ij})^\top
    \]
is the spatial distribution across all $I$ provinces in year $j$.

\textbf{Spatial consistency check} For each year $j$, we quantify how similar its spatial pattern $N_j$ is to that of a reliable baseline year $N_y$ by computing the coefficient of determination ($R^2$) between $N_j$ and $N_y$. A high $R^2$ indicates that year $j$ reproduces the benchmark’s spatial structure well, suggesting high data quality with low noise. Conversely, a low $R^2$ indicates a substantial deviation from the benchmark pattern, suggesting lower data quality.

The quantitative results of this spatial stability evaluation are presented in Table~\ref{tb:spatial_stability}. We first establish a benchmark using the original NPP-VIIRS satellite record (2013–2018). As these data show, the spatial relationship between modern NTL and the proxies is highly stable, with coefficients of determination ($R^2$) consistently high for both GDP ($R^2$ range: 0.930–0.983) and POP ($R^2$ range: 0.932–0.988). This stable range provides a standard for assessing the quantity of our reconstructed data.

\begin{table}[h]
\centering
\caption{\textbf{Quantitative evaluation of spatial stability.} EVAL uses 1992 as the baseline year, while NPP-VIIRS uses 2012 as the baseline year.}
\label{tb:spatial_stability}
\begin{tabular}{ccc|ccc|ccc}
\hline
\multicolumn{6}{c|}{EVAL} & \multicolumn{3}{c}{NPP-VIIRS} \\
\hline
YEAR & GDP & POP & YEAR & GDP & POP & YEAR & GDP & POP \\
\hline
1991 & 0.9609 & 0.9869 & 1993 & 0.9679 & 0.9881 & 2013 & 0.9827 & 0.9883 \\
1990 & 0.9486 & 0.9911 & 1994 & 0.9417 & 0.9718 & 2014 & 0.9633 & 0.9625 \\
1989 & 0.9385 & 0.9886 & 1995 & 0.9508 & 0.9683 & 2015 & 0.9702 & 0.9610 \\
1988 & 0.9212 & 0.9782 & 1996 & 0.8736 & 0.9506 & 2016 & 0.9490 & 0.9551 \\
1987 & 0.9103 & 0.9878 & 1997 & 0.8490 & 0.9322 & 2017 & 0.9477 & 0.9552 \\
1986 & 0.8732 & 0.9860 & 1998 & 0.8523 & 0.9594 & 2018 & 0.9297 & 0.9323 \\
\hline
\end{tabular}
\end{table}

Against this benchmark, our analysis of the EVAL dataset yields two key insights. First, EVAL demonstrates high internal consistency. The stability metrics for the 1986–1991 period (derived from PANDA-China) are shown to be highly comparable to those from the 1992–1998 period (derived from Li's NTL). As seen in Table~\ref{tb:spatial_stability}, the $R^2$ ranges for both GDP and POP are commensurate across the two periods. This indicates that the data source transition in 1992 did not introduce a systemic shock to the dataset's reliability. More critically, the stability of the reconstructed 1986–1991 EVAL data is found to be directly comparable to the NPP-VIIRS standard. The $R^2$ metrics for EVAL in this earliest period are largely on par with the stability range observed in the actual NPP-VIIRS record. This result provides strong quantitative evidence for the consistency and reliability of the EVAL dataset, validating the quality of the data generated for the contested 1986–1991 period.

\subsection{Validation of time-series consistency}
\begin{figure}[h] 
  \centering
  \includegraphics[width=0.7\textwidth]{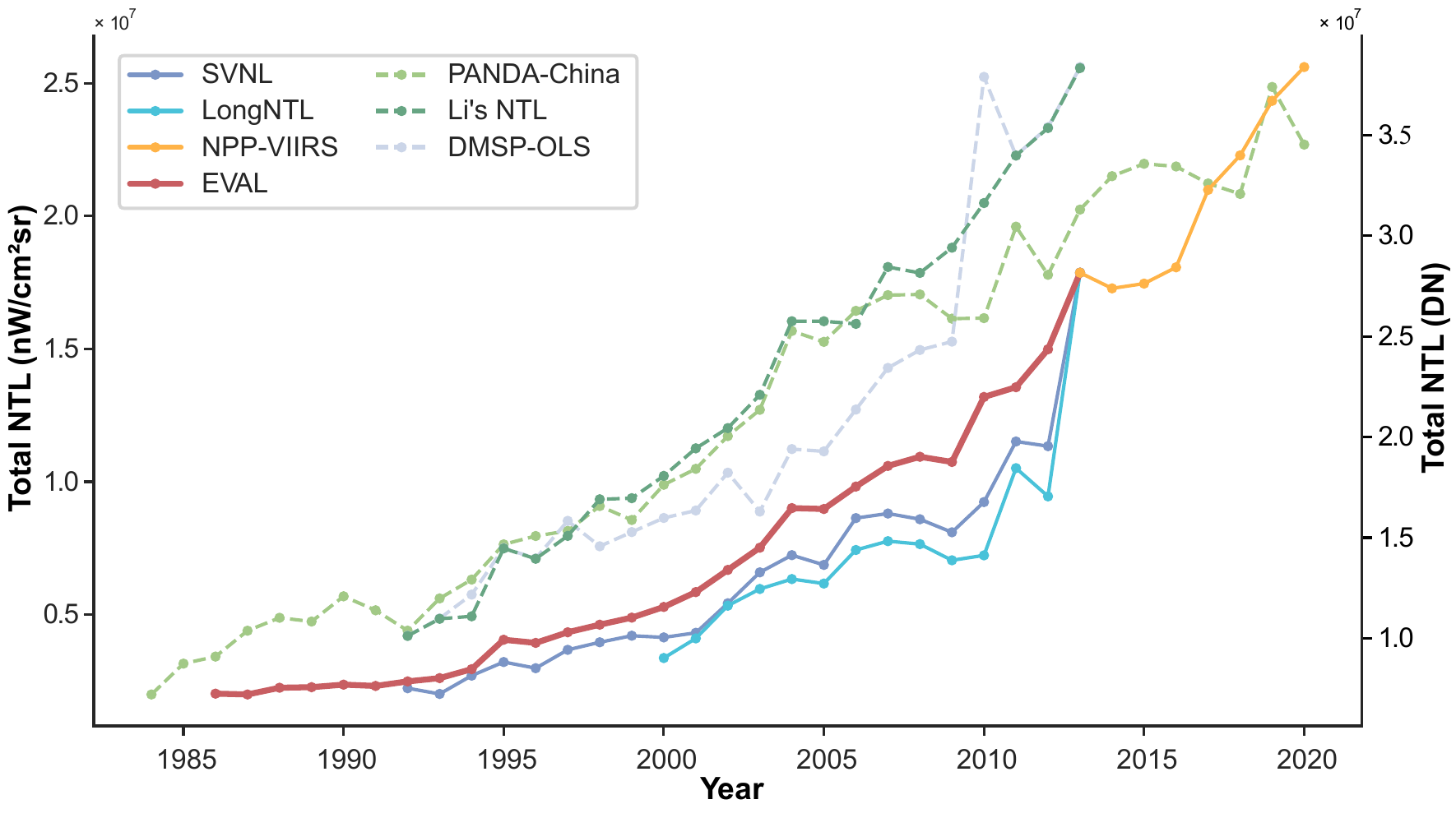}
  \caption{Comparison of Annual NTL Sums from different datasets.
EVAL, SVNL, LongNTL, and NPP-VIIRS use the left y-axis (Total NTL in $nW/cm^2\cdot sr$), while PANDA-China and DMSP-OLS use the right y-axis (Total NTL in Digital Number).}
  \label{fig6}
\end{figure}

To assess the temporal consistency of the EVAL, this study conducted a comparative analysis with several established long-term NTL datasets in Figure. \ref{fig6}. We statistically analyzed and compared the trends in the total annual NTL values for China across these datasets to evaluate EVAL's overall coherence and relative performance in capturing long-term light dynamics.
The analysis reveals that the time series of EVAL is only marginally shorter than that of the PANDA-China dataset, which is based on DMSP-OLS NTL data. When compared to other NPP-VIIRS-like NTL datasets, EVAL demonstrates a significant advantage in temporal coverage, offering unique support for high-precision, long-term studies. Furthermore, EVAL exhibits a high degree of temporal consistency across its entire time range, characterized by a stable growth trend. From 1992 to 2013, the trend in EVAL closely aligns with that of the corrected DMSP-OLS NTL datasets. Additionally, EVAL’s trend of variation is similar to that of existing VIIRS-like NTL products, but its values are significantly higher. This results in a stronger consistency with the official NPP-VIIRS NTL data, which in turn avoids the abrupt increase seen in LongNTL and SVNL during the 2012–2013 transition to the NPP-VIIRS record. 

\begin{figure}[h] 
  \centering
  \includegraphics[width=0.7\textwidth]{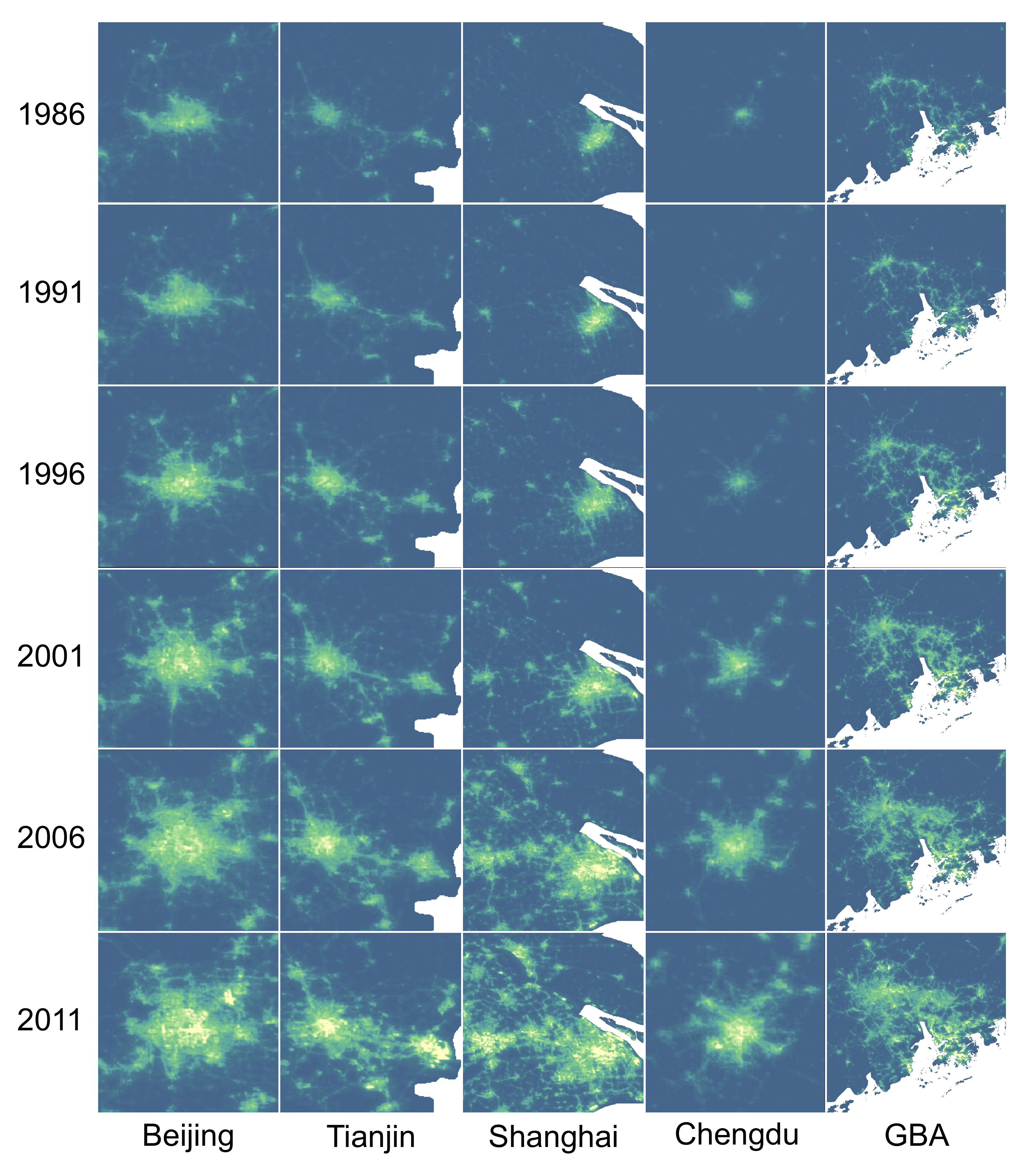}
  \caption{Time-series visualization of urban expansion in five urban agglomerations from 1986 to 2011. GBA denotes the Guangdong-Hong Kong-Macao Greater Bay Area.}
  \label{fig7}
\end{figure}

Leveraging its extended time series and high observational accuracy, EVAL could reconstruct the spatial and intensity patterns of nighttime lights from 1986 to the present. Figure. \ref{fig7} illustrates these dynamic changes across several major urban regions.

The result shows a clear and consistent trend of urban expansion and intensification over the 25-year period. In 1986, the nighttime lights in all depicted cities, were confined to small, distinct urban cores. By the mid-1990s, these cores had brightened and expanded significantly. From 2001 to 2011, the growth accelerated dramatically. This resulted in the formation of large, sprawling metropolitan areas and the coalescence of neighboring cities into vast, illuminated urban agglomerations, most notably in the Greater Bay Area (GBA) and the region encompassing Shanghai. The EVAL data effectively captures this transition from monocentric cities to polycentric urban clusters, detailing the trajectory of China's rapid urbanization.

\subsection{Socioeconomic correlation analysis}
Correlation analysis between NTL data and socioeconomic indicators(SEI) is a vital method for evaluating its quality and utility. A strong correlation indicates that the data accurately reflects socioeconomic activities, providing substantial value for research in fields like economic monitoring and regional development. We therefore compared VIIRS-like products against two key socioeconomic indicators: Gross Domestic Product (GDP) and Resident Population (POP) in Table. \ref{tb4}. We selected the 2000–2012 period for this analysis, the overlapping timeframe where all seven datasets were available. Among the evaluated products, DMSP-OLS, as the original NTL data, serves as a baseline for performance, showing an average correlation of 0.9356 with socioeconomic indicators. Li's NTL demonstrates an improvement upon this baseline through fine-tuning, achieving an average correlation of 0.9467.

Building upon Li's NTL, the reconstructed EVAL dataset demonstrates a comprehensive enhancement in performance. It surpasses its predecessor across every metric, registering superior correlation coefficients for GDP (0.9681), POP (0.9402), and the overall average (0.9542). This result confirms the effectiveness of our proposed reconstruction framework.

Furthermore, EVAL's superior performance extends to a broader comparison with other contemporary 500-meter resolution NTL products. The analysis reveals that EVAL consistently leads this category. Its correlation coefficients for both GDP and POP  are the highest among all 500m products, outperforming Tongji-NTL (0.8649, 0.8352), LongNTL (0.8622, 0.8846), and SVNL (0.8997, 0.9271). This consistent leadership across all metrics underscores EVAL's position as the most accurate and reliable 500m resolution NTL product in this study for socioeconomic applications.

\begin{table}[h]
	\centering
        \caption{Correlation Analysis of NTL Products with Socioeconomic Indicators.}
        \label{tb4}
	\begin{tabular}{c|ccccccc}
        \hline
		SEI & DMSP-OLS & Li's NTL &  Tongji-NTL & LongNTL & SVNL & EVAL\\ \hline
		GDP & 0.9475 & \uline{0.9560} & 0.8649 & 0.8622 & 0.8997 & \textbf{0.9681}\\ [0.3em]
            POP & 0.9236 & \uline{0.9374} & 0.8352 & 0.8846 & 0.9271 & \textbf{0.9402}\\ [0.3em]
            AVG & 0.9356 & \uline{0.9467} & 0.8506 & 0.8734 & 0.9134 & \textbf{0.9542}\\ [0.3em]
        \hline
	\end{tabular}
\end{table}

 We further leverage EVAL's unique long-term nature to conduct a historical correlation analysis spanning from 1986. This segmented analysis, shown in Table \ref{tb5}, reveals how the relationship between NTL and socioeconomic activity has evolved over nearly three decades.

The correlation was already strong in the initial 1986-1991 period, showing an average value of 0.8943. This relationship then intensified significantly during the 1992-1999 phase of rapid economic growth, where the average correlation reached a peak of 0.9632. This high level of correlation was maintained through the 2000-2012 period (average of 0.9542), reinforcing the findings from our previous comparison.

Overall, for the entire 25-year span from 1986 to 2012, the EVAL dataset maintained a remarkable average correlation of 0.9216 with socioeconomic indicators and 0.9673 with GDP. This confirms not only EVAL's superior performance against other products but also its robustness and reliability for long-term historical studies.

\begin{table}[h]
	\centering
        \caption{Correlation Coefficients between the EVAL Dataset and Socioeconomic Indicators, 1986–2012}
        \label{tb5}
	\begin{tabular}{c|cccc}
        \hline
		SEI & 1986-1991 & 1992-1999 & 2000-2012 & 1986-2012\\ \hline
		GDP & 0.9003 & 0.9791 & 0.9681 & 0.9673\\ [0.3em]
            POP & 0.8882 & 0.9472 & 0.9402 & 0.8758\\ [0.3em]
            AVG & 0.8943 & 0.9632 & 0.9542 & 0.9216\\ [0.3em]
        \hline
	\end{tabular}
\end{table}

\subsection{Ablation study}
To precisely evaluate the actual efficacy and contribution of each component within our proposed framework, we designed a series of ablation studies. Our model architecture, built upon a U-Net foundation, integrates SRF, MA and DFR. Its core objective is to overcome the prevalent issues of underestimation and omission of target details found in current methods. We conducted these studies by systematically removing each novel module to validate its effectiveness and contribution, with the results summarized in Table. \ref{tb6}.
\begin{table}[h]
	\centering
        \caption{Ablation results on the pipeline, including Structure Residual Fusion module, Multiscale Aggregator and Dual Feature Refiner.}
        \label{tb6}
	\begin{tabular}{@{}ccccccc@{}}
        \toprule
		\textbf{SRF} & \textbf{MA} & \textbf{DFR} & $\text{R}^2$\textbf{$\uparrow$} & RMSE\textbf{$\downarrow$} & PSNR\textbf{$\uparrow$} & UIQI\textbf{$\uparrow$} \\ \midrule
		× & - & × & 0.7924  & 1.0331 & 46.2399  & 0.8749\\ [0.3em]
            - & × & × & 0.7832 & 1.0612 & 46.2059  & 0.8584 \\ [0.3em]
            - & - & × & 0.7980 & 1.0242 & 46.5137  & 0.8783 \\ [0.3em]
            - & - & - & \textbf{0.8088} & \textbf{0.9965} & \textbf{46.7525} & \textbf{0.8968} \\ [0.3em]
        \bottomrule
	\end{tabular}
\end{table}

\paragraph{Analysis of the decoder}
We dissected the contributions of the two core modules within the decoder:

The inclusion of the SRF module improved the R² by 0.0056 and decreased the RMSE by 0.0089. While this numerical contribution is modest compared to the MA module, the SRF plays a crucial qualitative role. It guides the network to focus on key structural information and optimize the prediction of fine details, thereby proving its significant design value.Conversely, the MA module is the primary source of performance gain in the decoder. Removing it causes a sharp decline in model performance: the R² decreases by 0.0148, the RMSE increases by 0.0370, and the UIQI drops significantly by 0.0199. This result demonstrates that the MA effectively aggregates multi-level features and substantially enhances the model's overall performance.
\paragraph{Analysis of the refiner}

We conducted a holistic ablation analysis on the DFR module located at the end of the model to assess its necessity as a post-processing optimization step. We directly compared the full model equipped with the DFR against its counterpart without the module.

The results show that the introduction of the DFR module brought comprehensive performance enhancements: the R² increased by 0.0108, surpassing the 0.8 mark for the first time, while the RMSE correspondingly decreased by 0.0277, reaching a new minimum of 0.9965. Particularly noteworthy is the significant increase of 0.0185 in the UIQI, which directly reflects the high consistency of the predicted results with the ground truth labels in terms of structure, luminance, and contrast. This proves that the DFR module effectively performs a secondary refinement on the feature maps output by the decoder, significantly improving the visual quality and structural fidelity of the final output, and thereby further addressing the structral omission this study aims to solve.

\section{Data availability}
The Extended VIIRS-like Artificial Nighttime Light (EVAL) dataset generated in this study is available at the National Tibetan Plateau Data Center (TPDC) (\url{https://doi.org/10.11888/HumanNat.tpdc.302930}\cite{tian2025extended}).

\section{Code availability}
The programs used to generate all the results were Python 3.12. The code and scripts used for training, testing, and predicting the NTL data are available in the open GitHub repository “\url{https://github.com/Rookie1miao/EVAL_framework}”.

\section*{Acknowledgements}
This study is supported by National Key Research and Development Program of China (2022YFB3903703).

\section*{Author contributions}
  Yihe Tian: Conceptualization, Methodology, Software, Validation, Formal analysis, Visualization, Writing Kwan Man Cheng: Conceptualization, Methodology, Resources, Validation, Formal analysis Zhengbo Zhang: Resources, Validation Tao Zhang: Review \& Editing Zhehao Ren: Review \& Editing Junning Feng: Review \& Editing Suju Li: Review \& Editing Dongmei Yan: Review \& Editing Bing Xu: Review, Editing, Formal analysis, Funding acquisition, Supervision.
  
\section*{Competing interests}
The authors declare no competing interests.

\end{CJK*}
\bibliographystyle{unsrt} 
\bibliographystyle{plain}

\end{document}